\newif\ifarxivmode
\arxivmodefalse
\def\arxiv{\arxivmodetrue}

\arxiv

\ifarxivmode
	\documentclass{article} %
	\usepackage{natbib}
	\usepackage{authblk}
\else
	\documentclass{article} %
	\usepackage{iclr2019_conference,times}
	\input{math_commands.tex}
\fi
\usepackage{hyperref}
\usepackage{url}

\usepackage{times}
\usepackage{epsfig}
\usepackage{graphicx}
\usepackage{amsmath}
\usepackage{amssymb}
\usepackage{enumitem}

\usepackage[latin1]{inputenc}
\usepackage{subfig}
\usepackage{xspace}
\usepackage{booktabs}
\usepackage{color}
\usepackage{array,multirow}
\usepackage{listings}
\usepackage[title]{appendix}

\title{Déjà Vu: an empirical evaluation of the memorization properties of ConvNets \vspace{-8pt}
}

\ifarxivmode
	\author[$\dagger$,$\star$]{Alexandre Sablayrolles}
	\author[$\dagger$]{Matthijs Douze}
	\author[$\star$]{Cordelia Schmid}
	\author[$\dagger$]{Herv\'e J\'egou}

	\affil[ ]{\vspace{-3pt}$^\dagger$Facebook AI Research\; \quad \quad \quad $^\star$Inria}
	\usepackage{palatino}
\else
\fi

\begin{document}

\maketitle

\def \ie {\emph{i.e.}\xspace}
\def \eg {\emph{e.g.}\xspace}
\def \etal {\emph{et al.}\xspace}
\def \threshold {MAT\xspace}

\definecolor{darkgreen}{RGB}{0, 140, 0}

\ifarxivmode
	\newcommand{\rv}[1]{{\color{darkgreen}[\textbf{Rv}:#1]}}
	\newcommand \R {\mathbb{R}}
    \newlength{\tabcaplen}
    \setlength{\tabcaplen}{\linewidth}
    \newlength{\tabcaplenalt}
    \setlength{\tabcaplenalt}{\linewidth}
	\newcommand {\mypar}[1]{\medskip \noindent \textbf{#1} \quad}
\else
	\renewcommand{\rv}[1]{{\color{darkgreen}[\textbf{Rv}:#1]}}
	\renewcommand \R {\mathbb{R}}
	\newlength{\tabcaplen}
	\setlength{\tabcaplen}{0.4\linewidth}
	\newlength{\tabcaplenalt}
	\setlength{\tabcaplenalt}{0.52\linewidth}
	\newcommand {\mypar}[1]{\noindent \textbf{#1} \quad}
\fi

\renewcommand{\rv}[1]{{\color{darkgreen}[\textbf{Rv}:#1]}}
\newcommand{\alex}[1]{{\color{red}[\textbf{Alex}:#1]}}
\newcommand{\matthijs}[1]{{\color{blue}[\textbf{Matthijs}:#1]}}
\newcommand{\cordelia}[1]{{\color{magenta}[\textbf{Cordelia}:#1]}}

\newcommand \prob {\mathbb{P}}
\newcommand \D {\mathcal D}

\newcommand \positives {\mathcal P}
\newcommand \negatives {\mathcal N}

\newcommand \convnet {ConvNet\xspace}
\newcommand \convnets {ConvNets\xspace}
\newcommand \imsmall {Imagenet-1K\xspace}

\newcommand \private {\textit{private}\xspace}
\newcommand \public {\textit{public}\xspace}
\newcommand{\equival}[1]{\underset{#1}{\approx}}
\newcommand{\bigo}{\mathcal{O}}
\newcommand \myfig[1] {\includegraphics[width=\linewidth,trim={0 19.6cm 0cm 0},clip]{#1}}

\def \imnet       {\textit{Imnet1k}\xspace}
\def \imnettrain  {\textit{Imnet1k-train}\xspace}
\def \imnetval    {\textit{Imnet1k-val}\xspace}
\def \imnetbig    {\textit{Imnet22k}\xspace}
\def \yfcc        {\textit{Yfcc100M}\xspace}
\def \cifar       {\textit{CIFAR10}\xspace}
\def \tiny        {\textit{Tiny}\xspace}

\begin{abstract}
Convolutional neural networks memorize part of their training data, which is why strategies such as data augmentation and drop-out are employed to mitigate overfitting. 
This paper considers the related question of ``membership inference'', where the goal is to determine if an image was used during training. 
We consider it under three complementary angles. We show how to detect which dataset was used to train a model, and in particular whether some validation images were used at train time.
We then analyze explicit memorization and extend classical random label experiments to the problem of learning a model that predicts if an image belongs to an arbitrary set.
Finally, we propose a new approach to infer membership when a few of the top layers are not available or have been fine-tuned, and show that lower layers still carry information about the training samples.
To support our findings, we conduct large-scale experiments on Imagenet and subsets of YFCC-100M with modern architectures such as VGG and Resnet.

\end{abstract}

\section{Introduction}

The widespread adoption of convolutional neural networks~\citep{lecun1990handwritten} (\convnets) for most recognition tasks, was triggered by the advance of~\cite{krizhevsky2012imagenet} in image classification and subsequent deep architectures~\citep{simonyan15vgg,he16resnet}. 
Several works have analyzed these architectures from different perspectives. 
\cite{zeiler2014visualizing} have proposed DeconvNet to vizualize filter activations.  \cite{lenc2015understanding} analyze their equivariance.
\cite{mahendran2015understanding}  show how to invert them and synthetize images maximizing the response of different classes. 
\cite{ulyanov2017deep} analyze the image priors implicitly defined by \convnets. 

All these works increase our understanding of \convnets, but the complex issue of overfitting and its relationship to optimization are still not fully understood. 
Several strategies are routinely used to avoid overfitting, such as $\ell_2$-regularization through weight decay~\citep{ref_weight_decay91}, dropout~\citep{srivastava2014dropout}, and importantly, data augmentation~\citep{ada_arxiv17,debi2017cutpaste,paulin_data14}. 
Yet few works~\citep{zhang16understanding,yeom18privacyrisk} have analyzed the interplay of overfitting and memorization of training images in high-capacity classification architectures. 
Specifically, we are not aware of such an analysis for a modern \convnet such as ResNet-101 learned on Imagenet. %

In this paper, we consider the privacy issue of membership inference, \ie, we aim at determining if a specific image or group of images was used to train a model.
This question is important to protect both the privacy and intellectual property associated with images. 
For \convnets, the privacy issue was recently considered by \cite{yeom18privacyrisk} for the small MNIST and CIFAR datasets. The authors evidence the close relationship between overfitting and privacy of training images. 
This is reminiscent of prior membership inference attacks, which  employ the output of the classifier associated with a particular example to determine whether it was used during training or not~\citep{shokri16membershipinference}.  %
This is related to \cite{torralba2011unbiased}, who showed that a classifier can determine with high accuracy if an image comes from a dataset or another by exploiting the bias inherent to datasets. 
We discuss this relationship and show that we can detect whether a given network has been trained on some of the validation images.
This has a concrete application for machine-learning benchmarks: scores are often reported on a validation set with public labels, allowing a malicious or gawky competitor to artificially inflate the accuracy by training on validation images. 
Our test detects if it is the case, even if only part of the validation set is leaked to the training set.

We provide a qualitative upper bound on the capacity of popular convolutional networks to memorize a given number of images. 
More precisely, we construct a binary classifier as a drop-in replacement of the last layer, whose response is the membership test. 
Our tests carried out on VGG-16~\citep{simonyan15vgg} and ResNet-101~\citep{he16resnet} evidence different memorizing capabilities depending on the number of images and the amount of data augmentation (flip, cropping). %

Finally, we propose a new setting for membership inference that only considers intermediary layers of a network, thus extending membership inference to transferred and fine-tuned networks, that have become ubiquitous. 
Our membership inference does not require the last layer(s) of the original ConvNet to perform the test.  
This is important because, in many contexts, image recognition systems are built upon a trunk trained on a dataset and then fine-tuned for another task. 
Examples include Mask-RCNN~\citep{he2017mask} and models used for fine-grained recognition~\citep{hariharan2017low}. 
In both cases there are not enough training samples to train a full network: only the last layers of the networks are fine-tuned. 
In summary, our paper makes the following contributions: %
\begin{itemize}[leftmargin=8mm]
\item \vspace{-5pt}
 A simple statistical test to detect the ``signature" of a dataset in a trained convnet, and to detect if validation images where used to train the model (leakage). 
\item An empirical analysis of the explicit memorization capabilities of the ResNet and VGG architectures at a much larger scale than previously reported. We evaluate the factors impacting the memorization capabilities such as the number of images ``stored'' in the network and the equivariance hypotheses in data augmentation. 
\item A membership inference test that detects if an image was used to train the trunk of a network. To our knowledge, it is the first work on membership inference that attacks intermediate layers. 
\end{itemize}

\vspace{-5pt}
The paper is organized as follows. Section~\ref{sec:related} reviews related work. 
Section~\ref{sec:explicit} evaluates the capacity of \convnets to memorize a given set of images. 
Section~\ref{sec:dataset_level} considers the problem of determining if a particular dataset, \eg, the validation set, was used during training. %
Section~\ref{sec:implicit} focuses on detecting if a particular image has been used for training without accessing the network's output layer. 

\section{Related work \& datasets}
\label{sec:related}

Our work is related to the topics of overfitting and memorization capabilities of neural network architectures, which are able to perfectly discriminate random outputs in some cases~\citep{mackay2002informationtheory,zhang16understanding}. 
In the following, we distinguish \emph{explicit} from \emph{implicit} memorization (also called ``unintended memorization''~\citep{carlini18secret} in natural language processing systems). 

\mypar{Explicit memorization.} 
Neural network are capable of memorizing any random pattern. This property was analyzed in \cite{mackay2002informationtheory} for the single layer case. In MacKay's setup, the sender and receiver agree beforehand on a set of vectors $ (x_i)_{i=1}^n \in \R^d$. 
To transmit an arbitrary sequence of binary labels $y_1, \dots, y_n$, the sender learns a single-layer neural network that predicts the  $y_i$ from $x_i$, and sends its weights to the receiver. 
The receiver labels the points $x_1, \dots, x_n$ with the transmitted neural network to reconstruct the labels. %
The VC-dimension of this 1-layer model is $d$, so the model can fit perfectly as long as $ n \leq d$. 
MacKay extends this bound by showing that the sender can, with high probability, find a neural network fitting the output if $ n \leq 2d$, and that it is almost impossible to fit the model for $ n > 2d $.
The estimated capacity of this neural network is thus $2$ bits per parameter.

Determining the practical memorization capacity of ConvNets is not trivial. A few recent works~\citep{zhang16understanding,yeom18privacyrisk} evaluate how a network can fit random labels. 
\cite{zhang16understanding} replace true labels by random labels and show that popular ConvNets can perfectly fit them in simple cases, such as small datasets (CIFAR10) or Imagenet without data augmentation. 
\cite{krueger17memorization} extend their analysis and argue in particular that the effective capacity of ConvNets depends on the dataset considered. 
In a privacy context, \cite{yeom18privacyrisk} exploit this memorizing property to watermark networks. 
As a side note, random labeling and data augmentation have been used for the purpose of training a network without any annotated data~\citep{dosovitskiy2014discriminative,bojanowski2017unsupervised}.  
Our paper is also related to few works~\citep{kraska2017case,iscen2017memory} that learn indexes as an alternative to traditional  structures such as Bloom Filters or B-trees. 
In particular, \cite{kraska2017case} show that in some cases, neural nets outperform cache-optimized B-tree on real-world data. These works exploiting explicit memorization of neural networks are reminiscent of works~\citep{hopfield1982neural,personnaz1986collective,hinton1986distributed,plate1995holographic} on associative memories and, more generally, distributed representations.

\mypar{Implicit memorization and privacy risk in learning systems.} 
\cite{ateniese2015hacking} state: ``\emph{it is unsafe to release trained classifiers since valuable information about the training set can be extracted from them}''.
The problem that we address in this paper, \ie, to determine whether an image or dataset has been used for training, is related to the privacy implications of machine learning systems. They were discussed in the context of support vector machines~\citep{rubinstein09learning,Biggio14securitysvm}. 
In the context of differential privacy~\citep{dwork2006calibrating}, recent works~\citep{wang2016average,bassily2016algorithmic} suggest that guaranteeing privacy requires learning systems to generalize well, \ie, to not overfit. 
Note that there are systems providing differential privacy but that still leak information~\citep{ateniese2015hacking,balu14challenging}. 

\mypar{Membership Inference in images.} A few recent works~\citep{abadi16deep,hayes17logan,shokri16membershipinference,long18understanding} have addressed ``membership inference'' for images: determine whether an image has been used for training or not. 
\cite{yeom18privacyrisk} discuss how privacy, that can be broken by membership inference, is  connected to overfitting.
\cite{long18understanding} observe that some training images are more vulnerable than others and propose a strategy to identify them. %
\cite{hayes17logan} analyze privacy issues arising in generative models.
Most of these works were evaluated on small datasets like CIFAR10, or larger datasets but without data augmentation. %
Our work aims at being closer to realistic conditions. 
In the following, the analysis of a pre-trained network will be called ``attack'' performed by an ``attacker''.

\mypar{Dataset bias and inference.} 
\cite{torralba2011unbiased} evidence %
that a simple classifier can predict with high accuracy which dataset an image comes from. %
\cite{tommasi2017deeper} show that this bias still exists with ConvNets. %
In the next section of this paper, we re-visit this problem by proposing a \emph{dataset inference} method derived from an elementary membership inference test.

\mypar{Datasets used in our study.}
We use will several public image collections throughout our paper. 
\textbf{\imnet} refers to the subset of Imagenet~\citep{deng2009imagenet,russakovsky15imagenet} used during the ILSVRC-12 challenge. It consists of 1000~balanced classes, split in a training set (1.2M images) and a validation set (50k images). We use the regular split between train and val and denote them by \textbf{\imnettrain} and \textbf{\imnetval}, respectively. 
\textbf{\imnetbig} refers to the full Imagenet dataset.  %
It is built in the same way as \imnet, but with 21783~unbalanced classes. 
\textbf{\yfcc}~\citep{Thomee16} contains 99.2M~photos that have not been collected for image classification and thus are not representing specific classes or visual concepts.
\textbf{\tiny images} ~\citep{torralba2008} consists of 79M low-resolution images. 
\textbf{\cifar} is a subset of \tiny that has been labelled for image classification.
In our study, %
it is important that the dataset does not contain
duplicate images or images that overlap between the train and the test
set.  We have sanitized the datasets to avoid this problem using GIST descriptors %
and similarity search, %
see Appendix~\ref{sec:dedup} in the supplemental material for details.

\section{Explicit memorization -- network capacity}
\label{sec:explicit}

In this section, we explicitly train neural networks to memorize a given subset of images, so that it can  decide whether an image is in its memory or not at test time. 
We design a model $ f_{\theta} $ that distinguishes a set of \emph{in} images from \emph{out} images, where images unseen during training are \emph{out}. 

We repurpose the classification layer of standard models to output a binary label, depending on whether the image must be remembered or not. 
Our architecture plays an equivalent role to the discriminator in Generative Adversarial Networks (GAN): it needs to discriminate between positive and negative images. 
In our case, negative images are a large pool of images instead of the generated images in GANs. 
\cite{zhang16understanding} show that \convnets are able to overfit almost any random labelling of their input data, but in their experiment, the output for unseen images is undefined.

\subsection{Information-theoretic capacity}
\label{sec:capacitytheory}

We train our model to predict $1$ for a set of positive images $\positives$, and $0$ for all other (negative) images $ \negatives $. 
We assume that  $ \negatives $ represent a large image distribution, so $n=| \positives | \ll | \negatives | $, and we note $ N = | \positives | + | \negatives | $.
There are two notions of capacity relevant to our analysis: the capacity of a neural network, and the capacity needed to store a number $n$ of images among a larger set of size $N$.
We assume, following MacKay's analysis, that the capacity of a neural network is well approximated by its number of parameters. 
Assuming images have consecutive ids, it is sufficient to store the subset of ids of the positive images, which requires a capacity: 
\begin{equation}
C(n) = \log_2 {N \choose n} \approx n \log_2 \left( \frac{N}{n} \right) + \frac{n}{\log(2)},
\end{equation}
where the approximation holds for $ n \ll N$. 
This number scales almost linearly in the number of positive examples $n$, and logarithmically in the number of negatives. 

\subsection{Empirical analysis on tiny images}
\label{sec:tinynets}

\mypar{TinyNet.} We design a family of ConvNets with 4 convolutional layers and 2 fully-connected layers that take 32x32 images as input and output a binary classification.
There are 3 versions: TinyNet-1, (90k parameters), TinyNet-2 (300k parameters) and TinyNet-3 (2M parameters). 
Most parameters of these models are in the first fully connected layer, as in VGG (cf. Appendix \ref{app:tinynet}).

\mypar{Experimental setup.} 
We use a subset of $N=15M$ images from \tiny for these experiments.
We randomly sample $n$ images as positive examples, and treat the rest as negatives. 
At each epoch, we feed a random sample of negatives of the same size as the number of positives to the network. 
The reported accuracy is measured on a balanced set of positives and negatives. 
We consider four types of data augmentation:  ``none'', ``flip'' (random horizontal mirroring),  ``flip+crop$\pm$1'' (a random translation in $\{-1, 0, +1\}^2$), ``flip+crop$\pm$2''. %

\begin{figure*}[t]
{\scriptsize
\begin{tabular}{ccc}
TinyNet-1 & TinyNet-2 & TinyNet-3 \\
\includegraphics[clip,height=0.21\linewidth]{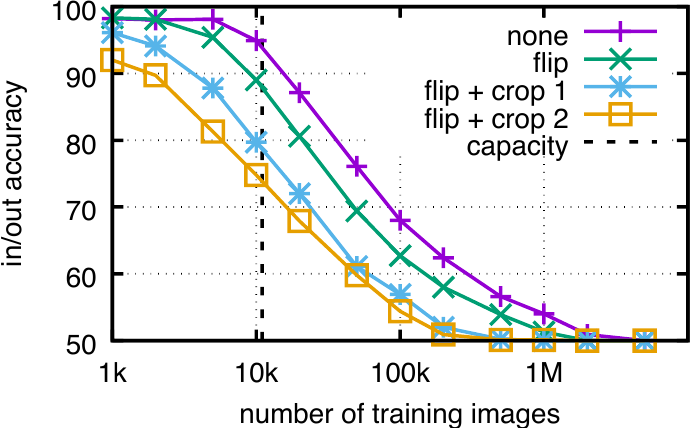} & 
\includegraphics[trim={1cm 0 0 0},clip,height=0.21\linewidth]{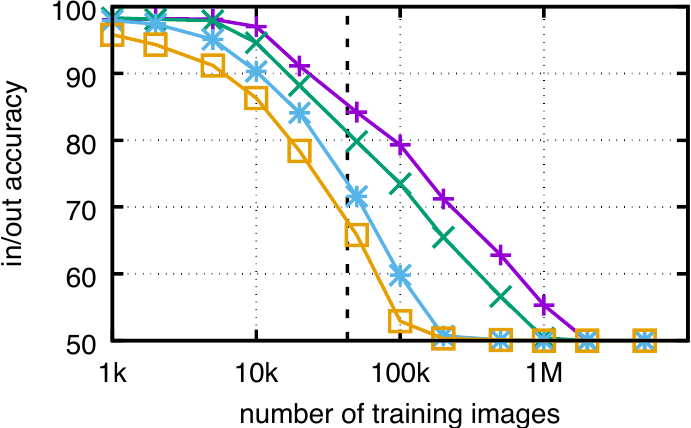} & 
\includegraphics[trim={1cm 0 0 0},clip,height=0.21\linewidth]{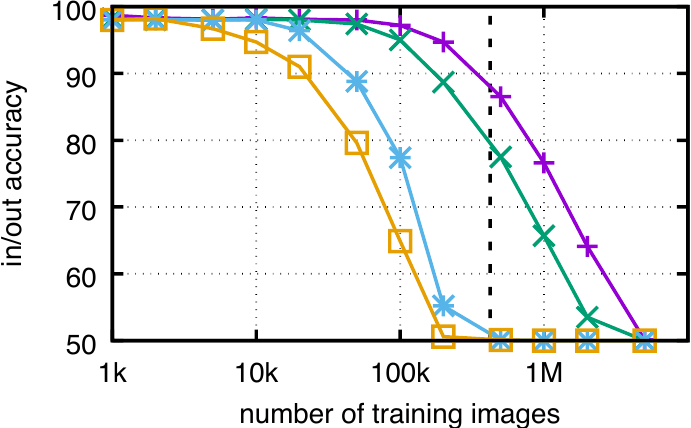} \\
\end{tabular}}
\caption{\label{fig:tinyacc}
In/out classification performance (train) on \tiny, for varying image subsets and number of images.
The colors indicate the type of data augmentation: purple=none, green=flip, cyan=flip+crop$\pm$1, orange=flip+crop$\pm$2. 
The vertical line shows the number of positive images $ n $ such that $ C(n) $ is the number of parameters of the network.
}
\end{figure*}

\mypar{Discussion.} 
Figure \ref{fig:tinyacc} shows the accuracy of the model as a function of the number of positive images for all TinyNets. 
Instead of a sharp drop between the over-capacity and the under-capacity regimes, we observe a smooth drop as the number of positives increases. Empirically, this transition phase happens  when the number of samples reaches the theoretical capacity of the network.

As expected, data augmentation reduces the memorization capacity of the network. 
For example, the accuracy of a network trained on $n$ images with flips is lower-bounded by the capacity of the same network trained on $2n$ images with no data augmentation. 
This lower bound is not tight, thanks to the generalization capability of the \convnet, which captures the patterns common to an image and its symmetric.
This generalization capability is obvious for stronger augmentations: for example with ``flip+crop$\pm$1'' TinyNet-2 can identify 10k images with 90\% accuracy, vs. 20k images without data augmentation, while this requires to treat 18 augmented versions of each image similarly.

\subsection{Experiments with large-scale architectures}

In this section, we extend the explicit memorization experiments to VGG-16, ResNet-18, and ResNet-101 networks with images coming from \yfcc. 
The capacity of these networks is much larger than in the tiny setting: Resnet-18 has 11.7M parameters and VGG-16 has 140M.

We set an initial learning rate of $10^{-2}$ and divide it by $10$ when the accuracy gets over 60\%, and again at 90\%.
We run experiments using either the center crop, or two data augmentations (flip, flip+crop$\pm5 $).
Figure~\ref{fig:convergenceflickr} shows convergence plots for several settings. 
Note, the x-axis is in epochs, that are 10$\times$ slower for $n=$100k images than $n=$10k images.
The longest experiment took 4 days on 4 GPUs .
VGG-16 and ResNet-101 converge at approximately the same number of epochs, irrespective of $n$. %
Data augmentation increases the number of epochs required to converge, eg. for the ResNets, flip up to twice more epochs to be trained. 
VGG is a more shallow and higher capacity network; in general it converges faster and it handles crops better than the ResNet variants. 

\renewcommand \myfig[1] {\includegraphics[clip,width=0.29\linewidth]{#1}}
\begin{figure*}[t]
\begin{tabular}{cccc}
& resnet-18 & resnet-101 & VGG-16 \\
\rotatebox{90}{\hspace*{1cm}$n$=10k} &
\myfig{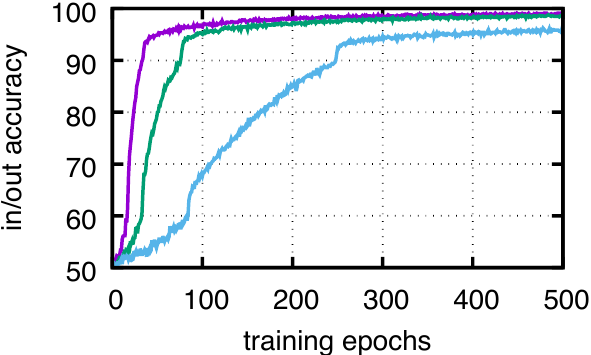} &
\myfig{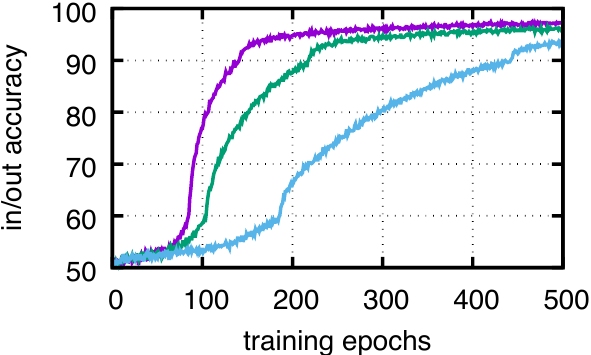} &
\myfig{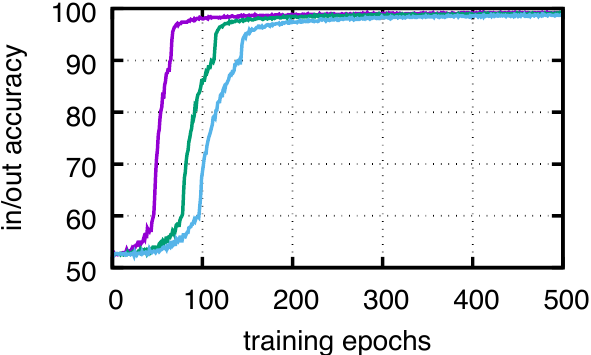} \\
\rotatebox{90}{\hspace*{1cm}$n$=100k} &
\myfig{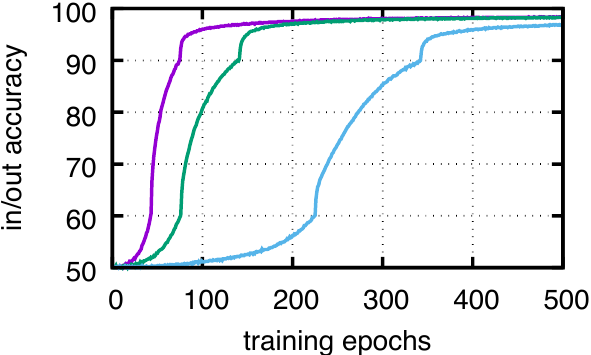} &
\myfig{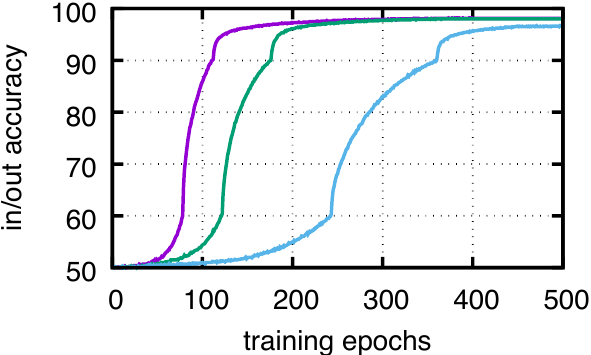} &
\myfig{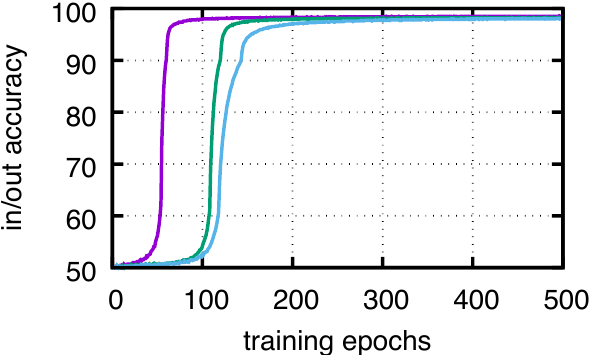} \\
\end{tabular}
\vspace{-3pt}
\caption{\label{fig:convergenceflickr}
	Accuracy over iterations of the in/out training on \yfcc for different networks and amount of data augmentation (indicated by color: purple= none, green = flip, cyan = flip+crop$\pm$5).
}
\vspace{-10pt}
\end{figure*}
\renewcommand \myfig[1] {\includegraphics[width=\linewidth,clip]{#1}}

The outcome of our analysis is that explicit memorization of a large amount of images is possible, albeit more difficult with data augmentation. 
In real use cases, the number of images that can be stored explicitly with perfect accuracy is practically much lower than the number of network parameters.
This set of experiments provides an approximate upper-bound for the problem of membership inference: if a given model cannot perfectly remember a set of images when trained to do so, it will likely not be able to remember all the images of the training set when trained for classification.
\section{Dataset detection and leakage} 

\label{sec:dataset_level}

In this section, we detect whether a group of samples or a dataset has been used to train a model. 
This problem encompasses the particular case of dataset bias~\citep{torralba2011unbiased} and is more difficult, as we need to distinguish datasets even if they share the same statistics, acquisition procedure and labelling process. 
For instance, we want to be able to determine if images from the validation set of \imnet were used at train time.

\begin{figure*}[t]
{\scriptsize
\includegraphics[width=0.48\linewidth]{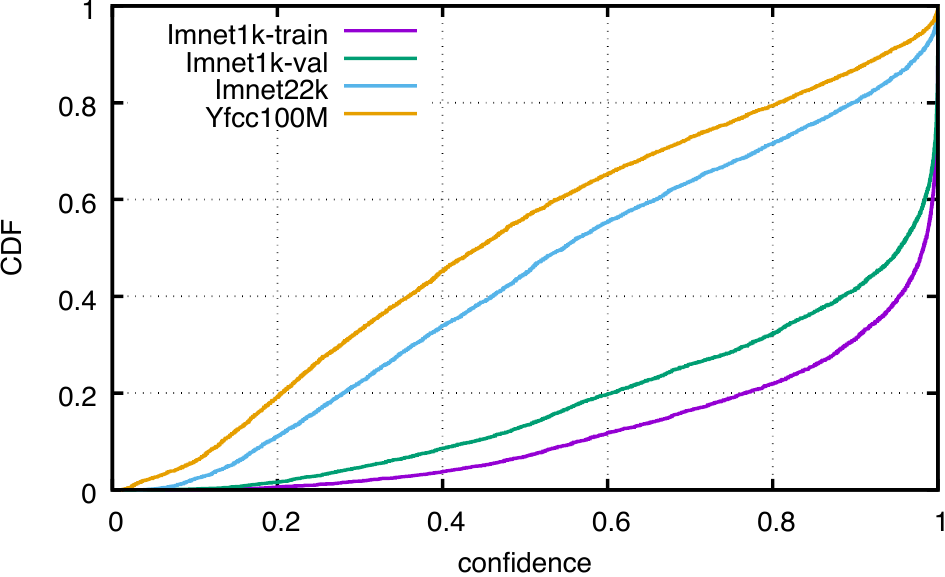}  \hfill
\includegraphics[width=0.48\linewidth]{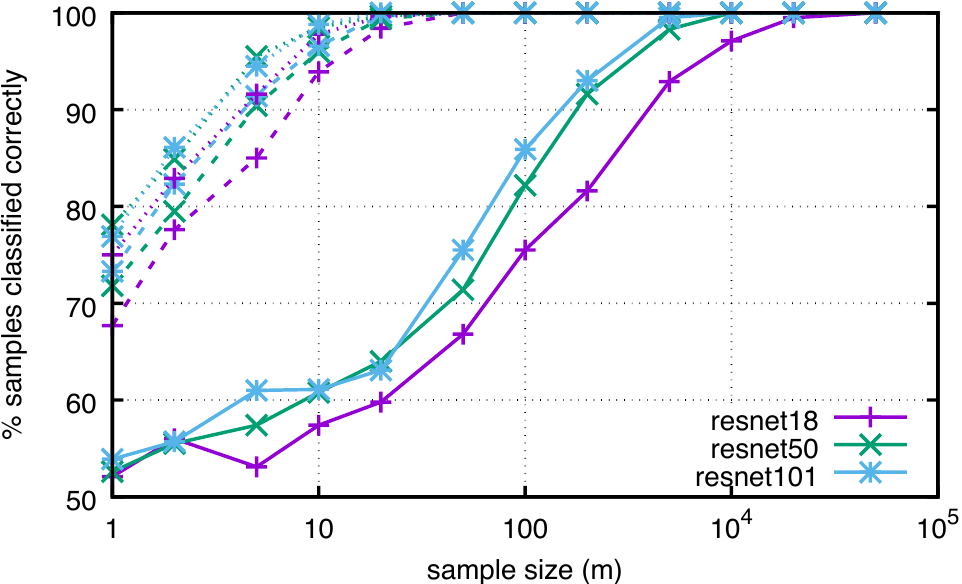} 
}
\caption{\label{fig:twodata}\label{fig:maxdistribution}
\emph{Left:} Cumulative distribution of the maximum classification score for a sample of 5000 images taken from 4 datasets. 
	\imnettrain served as the training set and therefore \imnet images (both train and val) have higher confidence.
\emph{Right:} binary classification accuracy (\%) of a sample of $m$ elements from the training set \imnettrain w.r.t. three other datasets: 
	 \imnetval (solid), \imnetbig (dashed) and \yfcc (lines). The architecture is indicated by the line color. 
}
\end{figure*}

\mypar{Hypothesis and problem statement.}
We assume that there are two \textit{data sources} $ S_1, S_2 $ and each source $ S_j $ yields samples $ x_1^{(j)}, x_2^{(j)}, \dots, x_{m}^{(j)} $. 
The attacker is given access to a model $ f_{\theta} $; in this paper, we assume $ f_{\theta}(x)$ is the maximal activation of the softmax layer, aka. the \textit{confidence} of the model.
The cumulative distribution of the confidence for a model trained on \imnettrain is shown in figure \ref{fig:maxdistribution}: most samples coming from the source \imnettrain have a very high confidence, while the distribution of the source \imnetval is more balanced and unrelated sources (\yfcc, \imnetbig) tend to have a more uniform distribution.

We consider two attack scenarii on the model $ f_{\theta}$. 
In the first scenario, we have a set of $ m $ samples that come from either $S_1$ or $S_2$ and we want to determine which source they come from. %
In the second scenario, we want to determine if the model has been trained with samples from a validation set, and thus look at whether the two source distributions corresponding to the validation and the test are different.

We compare confidence distributions using the Kolmogorov-Smirnov (K-S) distance. Given two cumulative distributions $ F $ and $ G $, the K-S distance is 
$%
d_{\text{KS}}(F, G) = \sup_{x} | F(x) - G(x)|. 
$ %
We use the K-S distance to determine if two distributions are similar.

\subsection{Confidence as a signature of a dataset}

In this section the samples $x_1, \dots, x_m$ come from either source $ S_1 $ or $ S_2 $.
The attacker uses the following decision rule: compute the K-S distance between $x_1, \dots, x_m$ and $ S_1$ (resp. $ S_2 $), and assign the samples to the closest source.

\mypar{Results and observations.} 
The results are reported in Figure~\ref{fig:twodata}. 
We can distinguish \imnettrain from \yfcc with very few (10-20) samples. More interestingly, the same number of samples allow us to separate \imnetbig from \imnettrain, and with 500 images we can distinguish \imnettrain from \imnetval. %
This shows that, even with a relatively low number of images, an attacker can determine that a given image collection was used for training. 
The figure also shows that the test is easier for networks with a higher capacity, that tend to overfit more.

\subsection{Detecting leakage}

We now assume that we are given a model for which we suspect that part of the validation set was used for training (\emph{leakage}). 
For a number of datasets (\eg, Imagenet, Pascal VOC), the labels of the validation set are publicly available, and models are often compared using validation accuracy.
A malicious person could train a model using the training set and part of the validation set, and then report validation accuracy to artificially inflate the performance of the model. 

The attack we propose is a two-sample K-S test to determine if leakage has occurred or not. 
We assume that no sample from the test set has leaked (labels are not public in most cases).
The null hypothesis of our test is that the validation and test sets have the same distribution.
We compute the K-S distance between the validation and test sets, and reject the null hypothesis if this distance is high.
The distance threshold $t$ is set such that the null hypothesis is incorrectly rejected with a low probability $\alpha$, corresponding to the $p$-value. 
For large samples, Smirnov's estimate of the threshold corresponding to a $p$-value of $ \alpha $ is~\citep{feller49}: 
\begin{equation}
t = c(\alpha) \sqrt{\frac{n + m}{nm}}
\textrm{\ \ \ where\ \ \ } 
c(\alpha) = \sqrt{-\frac{1}{2} \log \left( \frac{\alpha}{2} \right) }. 
\end{equation} 

We ran experiments on Imagenet using Resnet-18 and VGG-16, with $s \in \{1, 2, 5, 10, 20\}$ images per class of the validation set in addition to the training set to fit the model.
Table \ref{tab:leakage} reports the $p$-value of the different tests.
We can see that when $ 10 $ images per class are leaked, the K-S test predicts that leakage has happened with a very high significance.
When $ 5 $ images per class or less are used, we cannot reject the null hypothesis and thus cannot claim that leakage has happened.

\begin{table}[t]
\centering
\begin{minipage}{\tabcaplenalt}
\caption{\label{tab:leakage}
Kolmogorov-Smirnov tests on \imnet validation and test sets for various levels of leakage.
The K-S test provides a level of significance ($p$-value) rather than a yes/no answer. 
Lower values indicate high confidence that the validation and test sets distributions are different, hinting that leakage has occurred.
If only $1$ image per class of the validation set has leaked, we cannot conclude from this test that there has been leakage. 
Conversely, when $10$ images or more have leaked, we can conclude with high significance that leakage has occurred.
}
\end{minipage}
\hfill
\begin{tabular}{clll}
\toprule
Nb. of Images 	&   	& 	\\ \# per class leaked	&  Resnet-18	& VGG-16	\\ \midrule
1			&  $ 0.888 $	& $0.494$\\
2			&  $ 0.228$	& $0.107$\\
5			&  $ 0.068$	& $0.014$\\
10			&  $ < 10^{-4}$	& $ < 10^{-4}$\\
20			&  $ < 10^{-4}$	& $ < 10^{-4}$\\ \bottomrule
\end{tabular}
\end{table}

\section{Implicit memorization \& membership inference}
\label{sec:implicit}

This section tackles the more difficult problem of membership inference in trained models.
From a trained model and an image the attacker has to determine whether the image was used to train the model. 
In our new setting, upper layers are not available (due to \eg finetuning on a downstream task).
We provide baselines for VGG16 and Resnet and extend the traditional attacks to our  setup.

The literature~\citep{abadi17protection} distinguishes two cases types of membership inference: (1) all layers are available (\textit{all-layers}), (2) only the final output of the network is available (\textit{final-output}). 
There is currently no attack that performs substantially better in \textit{all-layers} than in \textit{final-output}.
This seems counter-intuitive but we confirmed it in  preliminary experiments. 
Our new setup, \textit{partial-layers}, is adapted to transfer learning: 
only a certain number of bottom layers are available for attack, the remaining layers were destroyed by retraining on an unrelated task.
This task is more difficult than \textit{all-layers} since it has less parameters available, and thus more difficult than \textit{final-output}.

\subsection{Evaluation protocol and baselines}
\label{sec:baselines}

We assume that there are three disjoint sources of data: a \emph{public} set, a \emph{private} set, and an \emph{evaluation} set.
A model is trained on the private set. 
The attacker has access to the lower layers of this model and to the public set. %
After the attack is carried out, the evaluation is ran on images from the private and evaluation sets. 

We divide \imnet equally into two splits (each with half of the images per class).
We hold out $ 50 $ images per class in the first split to form the evaluation set, and form the private set with the rest of this split.
The second split is used as the public dataset.
We conduct the membership inference test by comparing the prediction of the attack model on the private set and on the evaluation set. 
For this purpose, we consider the two baseline methods.

\mypar{Bayes rule.} 
A simplistic membership inference attack is to predict that an image comes from the training set if its class is predicted correctly, and from a held-out set otherwise. 
We note $p_{\text{train}} $ (resp. $ p_{\text{test}}$) the classification accuracy on the training (resp. held-out) set, and assume a balanced prior on membership. 
According to Bayes' rule, the accuracy of the heuristic is (see Appendix~\ref{sec:probaderiv} in the supplementary material for the derivation):
\begin{equation}
p_{\text{bayes}}  = 1/2 + (p_{\text{train}} - p_{\text{test}} )/2.
\end{equation}
Since $p_{\text{train}} \geq p_{\text{test}}$ this  heuristic is better than random guessing (accuracy $1/2$) and the improvement is proportional to the overfitting gap $p_{\text{train}} - p_{\text{test}}$.

\mypar{Maximum Accuracy Threshold (\threshold).} 
\label{sec:threshold_adv}
\cite{yeom18privacyrisk} propose an attack on the loss value: a sample is deemed part of the training set if its loss is below a threshold $ \tau $. %
If $ F_{\text{train}} $ (resp. $  F_{\text{heldout}} $) is the cdf of the loss on the train (resp. held out), the accuracy of the \threshold is:
\begin{equation}
p_{\text{threshold}} = \max_{\tau} ~1/2 + 1/2 \left(F_{\text{train}}(\tau) - F_{\text{heldout}}(\tau)\right)
\end{equation}
As $F_{\text{train}}(\tau) \geq F_{\text{heldout}}(\tau) $, this heuristic is also better than random guessing. 
In practice, $ \tau $ is estimated with samples or simulated by training models with known train/heldout split.

\subsection{Membership inference with a truncated network}

In this section, we provide a simple method to attack networks in the \textit{partial-layers} setting.
We use the available public data to retrain the missing layers, and apply either the Bayes or the \threshold attack, as if there was no fine-tuning at all. 
We found this method to be more accurate than another variant that we designed with shadow models~\citep{shokri16membershipinference}, as detailed in the supplemental material (Appendix~\ref{app:shadow}). %

\begin{table}[t]
\centering
\begin{minipage}{\tabcaplen}
\caption{\label{tab:attack_baselines}
Accuracy of membership inference attacks on the softmax layer of the models (\textit{final-output}). 
Data augmentation reduces the gap between the training accuracy and the held-out accuracy, thus decreasing the accuracy of the Bayes attack and the \threshold attack.
}
\end{minipage}
\hfill
\begin{tabular}{@{}llcc@{}}
\toprule
Model     		& Augmentation 		& Bayes baseline	& \threshold \\ \midrule
Resnet101 	& None				& 76.3			& 90.4 \\ 
			& Flip, Crop $\pm5$		& 69.5			& 77.4 \\ 
			& Flip, Crop 			& 65.4			& 68.0 \\ \midrule
VGG16     	& None   				& 77.4			& 90.8  \\
    	 		& Flip, Crop $\pm5$		& 71.3			& 79.5  \\
     			& Flip, Crop			& 63.8			& 64.3  \\ \bottomrule
\end{tabular}
\end{table}

\subsection{Experiments on large convnets}

\mypar{Classification models.}
We experiment with the popular VGG-16 \citep{simonyan15vgg} and Resnet-101 \citep{he16resnet} architectures.
The private model is learned in $ 90 $ epochs, with an initial learning rate of $ 0.01 $, divided by $ 10 $ every $ 30 $ epochs. 
Parameter optimization is conducted with stochastic gradient descent with a momentum of $ 0.9 $, a weight decay of $ 10^{-4} $, and a batch size of $ 256 $.
To assess the effect of data augmentation, we train different networks with varying data augmentation: flip+crop$\pm$5, flip+crop, flip+crop+resize, or none. %

\mypar{Attack models.} 
We evaluate both the bayes and \threshold methods to estimate the performance on \textit{final-output}. 
The results are shown in Table \ref{tab:attack_baselines}.
As we can see, it is possible to guess with a very high accuracy if a given image was used to train a model when there is no data augmentation.
Stronger data augmentation reduces the accuracy of the attacks, that still remain above $ 64\%$.

The results of our attack in the more challenging \textit{partial-layers} setting are shown in Table \ref{tab:attacks_partial}. 
We can see that even without the last layers, it is possible to infer training set membership of an image.
The attack performance depends on two factors: the layer at which the attack is conducted, and the data augmentation used to train the original network. 
As expected, it is more difficult to attack a network that has been trained with more data augmentation, or that has only lower layers available. 
More importantly, these experiments show that intermediary layers still carry out information about the images used for training the model.

\begin{table}[t]
\centering
\begin{minipage}{\tabcaplen}
\caption{\label{tab:attacks_partial}
Accuracy of membership inference attacks on intermediate layers of Resnet-101 and VGG-16 models (\textit{partial-layers}).  \newline
\textit{Last block} corresponds to the softmax, and respectively the fully connected layers (for VGG-16) and the 4-th stage of Resnet (for Resnet-101).
}
\end{minipage} 
\hfill
\begin{tabular}{lllcc}
\toprule
Augmentation 		& Truncate	& Resnet-101	& VGG-16 \\ \midrule
None			& Softmax			& 73.4 		& 74.8\\
				& Last block 		& 53.1 		& 51.7\\ \midrule
Flip, Crop$\pm5$	& Softmax			& 65.7 		& 67.3\\
				& Last block		& 53.1		& 52.2\\ \midrule
Flip, Crop 			& Softmax			& 60.8		& 58.5\\ 
				& Last block		& 52.9 		& 53.2\\ \bottomrule
\end{tabular}
\end{table}

\section{Conclusion}
\label{sec:conclusion}

We have investigated the memorization capabilities of neural networks from different perspectives. Our experiments show that state-of-the-art networks can remember a large number of images and distinguish them from unseen images. We have analyzed networks specifically trained to remember a set of images and the factors influencing their memorizing and convergence capabilities. 
It is possible to determine whether an image set was used at training time, even with full data augmentation. On the contrary, the accuracy of determining if a single image was used is low when considering full data augmentation on a large training set such as Imagenet. This implies that data augmentation is an effective privacy-preserving method. Our last contribution is a method that detects training images better than chance even with no access to the last layers, under limited data augmentation. 

\mypar{Final remark:} The curious reader may have noticed that our title echoes the one of a previous user study~\citep{dhamija2000deja}, in which the authors discussed the feasibility of authenticating humans by their capabilities to recognize a set of images.

\bibliography{egbib}
\bibliographystyle{iclr2019_conference}

\clearpage

\appendix
\begin{appendices}
\section{Probabilistic derivations}
\label{sec:probaderiv}

\subsection{Bayes attack} Let $ C $ denote the event that the prediction of the neural network is correct and $ S $ the random variable that indicates whether the sample comes from the training set. 
We therefore have:
\begin{align}
\prob(C=1|S=1) = p_{\text{train}},&~\prob(C=1|S=0) = p_{\text{test}} \\
\prob(S=1) = & \prob(S=0) = 1/2. 
\end{align}

The accuracy of Bayes attack is:
\begin{align}
\prob(C=S) &= \prob(C=1~|~S=1) \prob(S=1) \\ &+ \prob(C=0~|~S=0)\prob(S=0) \\
&= \frac{1}{2} (p_{\text{train}} + 1 - p_{\text{test}}).
\end{align}

\subsection{Equivalence between Kolmogorov-Smirnov and threshold attacks}

If we consider the particular case of a subset of $m=1$ image, 
we show in this section that the decision boundary induced by the K-S distance is the same as the \threshold described in Section~\ref{sec:threshold_adv}. 
Yet there are two significant differences between the K-S attack and the \threshold: we consider \textit{confidence} instead of the loss value, and the optimal threshold is computed differently. 
Our attacks with the K-S distance can therefore be seen as a generalization of the membership inference proposed by \cite{yeom18privacyrisk}.

We assume that we have two cumulative distributions $ F $ and $ G $  such that $ \forall x, F(x) \geq G(x) $. 
We want to show that the K-S rule is equivalent to a threshold rule.
Denoting by $ \delta_x $ the Dirac distribution centered on $x$, we have:

\begin{align}
&d_{\text{KS}}(\delta_x, F) \leq d_{\text{KS}}(\delta_x, G) \\
\iff &\frac{1}{2} - \large| F(x) - \frac{1}{2}|  \leq \frac{1}{2} - | G(x) - \frac{1}{2}| \\
\iff &| G(x) - \frac{1}{2}|  \leq | F(x) - \frac{1}{2}|. 
\end{align}
The two following cases are easy:
\begin{align}
G(x) \leq F(x) \leq 1/2 & \Rightarrow d_{\text{KS}}(\delta_x, F) \leq d_{\text{KS}}(\delta_x, G),   \label{eq:tau_left}\\
F(x) \geq G(x) \geq 1/2 &\Rightarrow d_{\text{KS}}(\delta_x, F) \geq d_{\text{KS}}(\delta_x, G).\label{eq:tau_right}
\end{align}

For the last case, the set $ I $ for which $ G(x) \leq  1/2 \leq F(x)$ is an interval. 
On this interval, $ | F(x) - 1/2|  - | G(x) - 1/2 | = F(x) + G(x) - 1 $.
$ F + G $ is increasing, and thus there exists a threshold $ \tau $ such that for $ x \in I$:
\begin{equation}
x \leq \tau \iff d_{\text{KS}}(\delta_x, F) \leq d_{\text{KS}}(\delta_x, G). \label{eq:tau_middle}
\end{equation}

With Equations \ref{eq:tau_left} and \ref{eq:tau_right}, Equation \ref{eq:tau_middle} extends to all $x$.

\section{De-duplicating the datasets}
\label{sec:dedup}

In this section, we describe the de-duplication processing applied to the datasets used in explicit memorization experiments. 
This process ensures that near-duplicate images do not get assigned different labels, and thus makes learning and evaluation more reliable.

\subsection{Description and matching of duplicates} 
\begin{figure}[t]
\begin{center}
\includegraphics[width=0.5\linewidth]{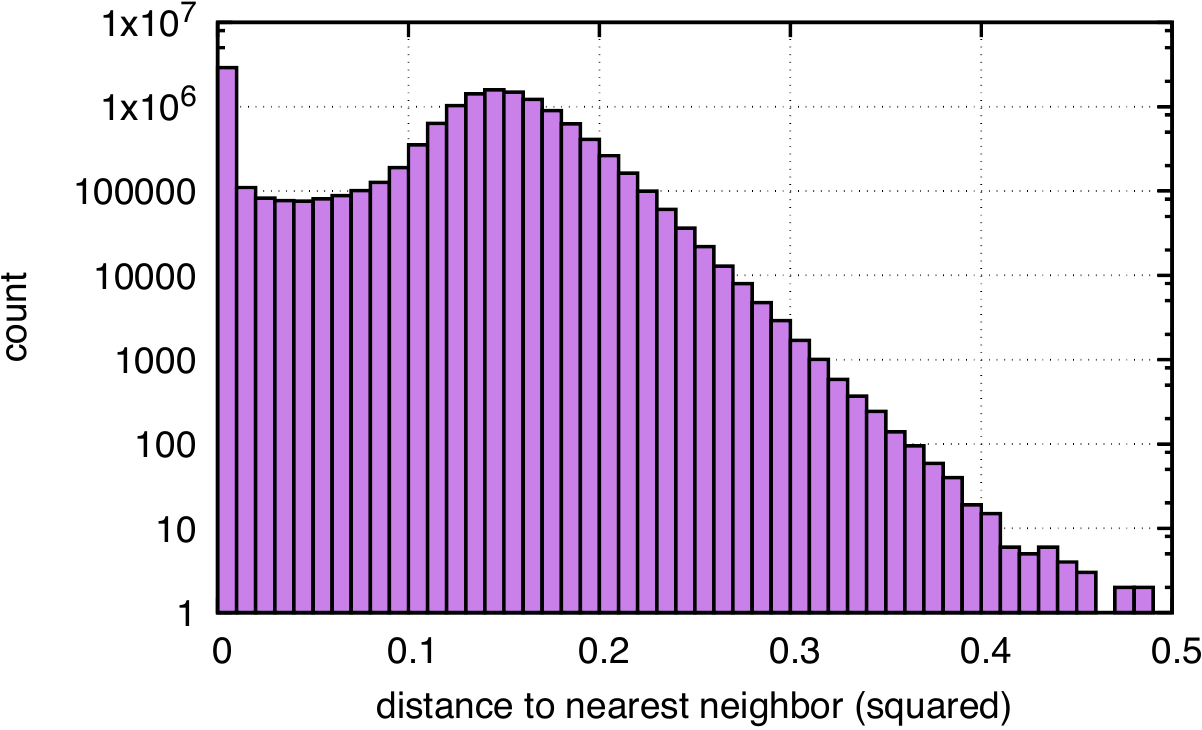}
\end{center}
\caption{\label{fig:distancetoNN}
	Histogram of distances of the images of \imnetbig to their nearest neighbor. 
}
\end{figure}

We compare images using GIST~\citep{oliva2006building}, a simple hand-crafted descriptor that was shown to perform well on moderate image transformations~\citep{douze2009gist}. 
We compute the approximate k-nearest neighbor graph on each dataset using Faiss~\citep{johnson2017billion}. 
Figure~\ref{fig:distancetoNN} shows the histogram of distances for the images of \imnetbig to their nearest neighbor:
the bin around $[0, 10^{-2}]$ contains more images than the following bin $[10^{-2}, 2.10^{-2}]$, which is due to duplicates in the dataset.

Images that are bit-wise exact are unambiguous duplicates -- in fact they are often already removed beforehand from the datasets because they are easy to detect by computing a hash value on the content.
Beyond this extreme case, the notion of ``duplicate'' is ambiguous: 
images that are re-encoded, resized, slightly cropped should be considered duplicates,  
but the case of larger transformations is less obvious (e.g., photos of the same painting, consecutive frames of a video).

\begin{figure*}[t]
\begin{minipage}{0.30\linewidth}
\centering TinyNet 1 \\ 
\myfig{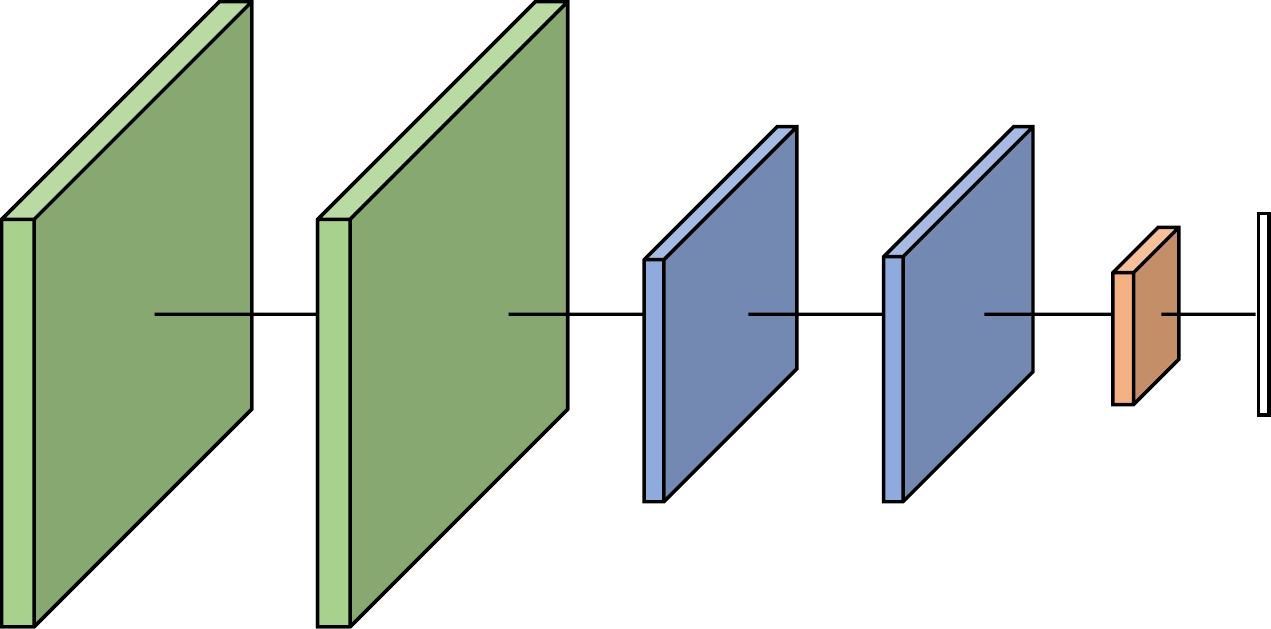} \\
\end{minipage}
\hfill
\begin{minipage}{0.30\linewidth}
\centering TinyNet 2 \\ 
\myfig{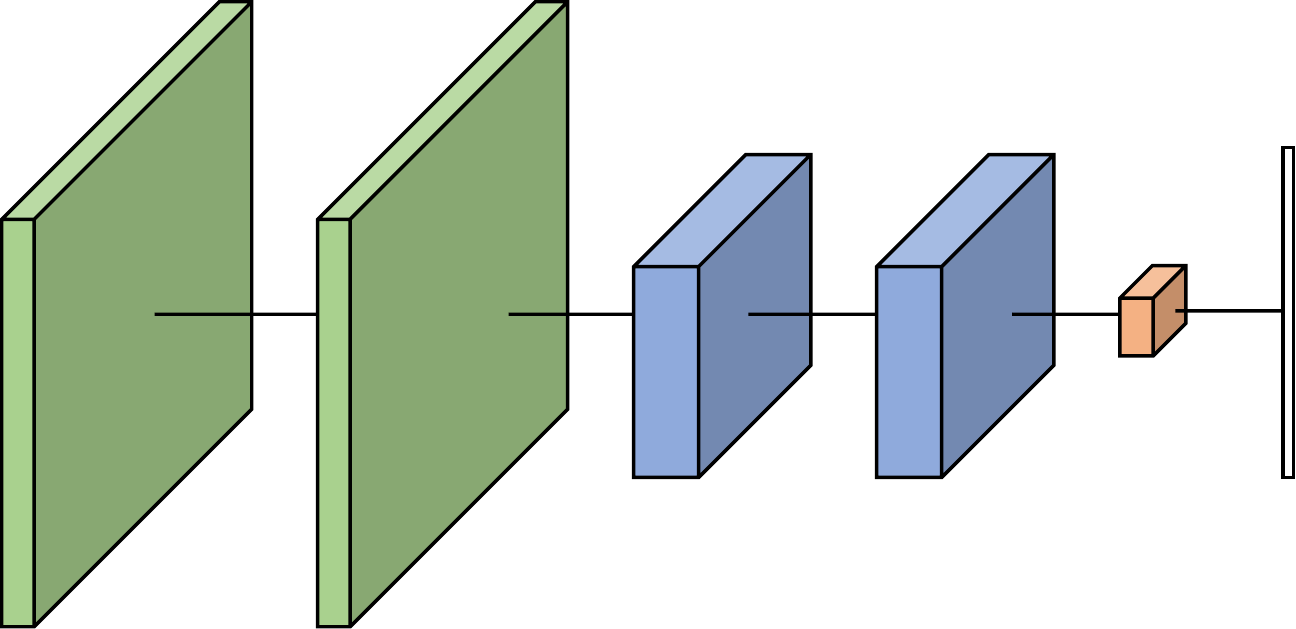} \\
\end{minipage}
\hfill
\begin{minipage}{0.30\linewidth}
\centering TinyNet 3 \\ 
\myfig{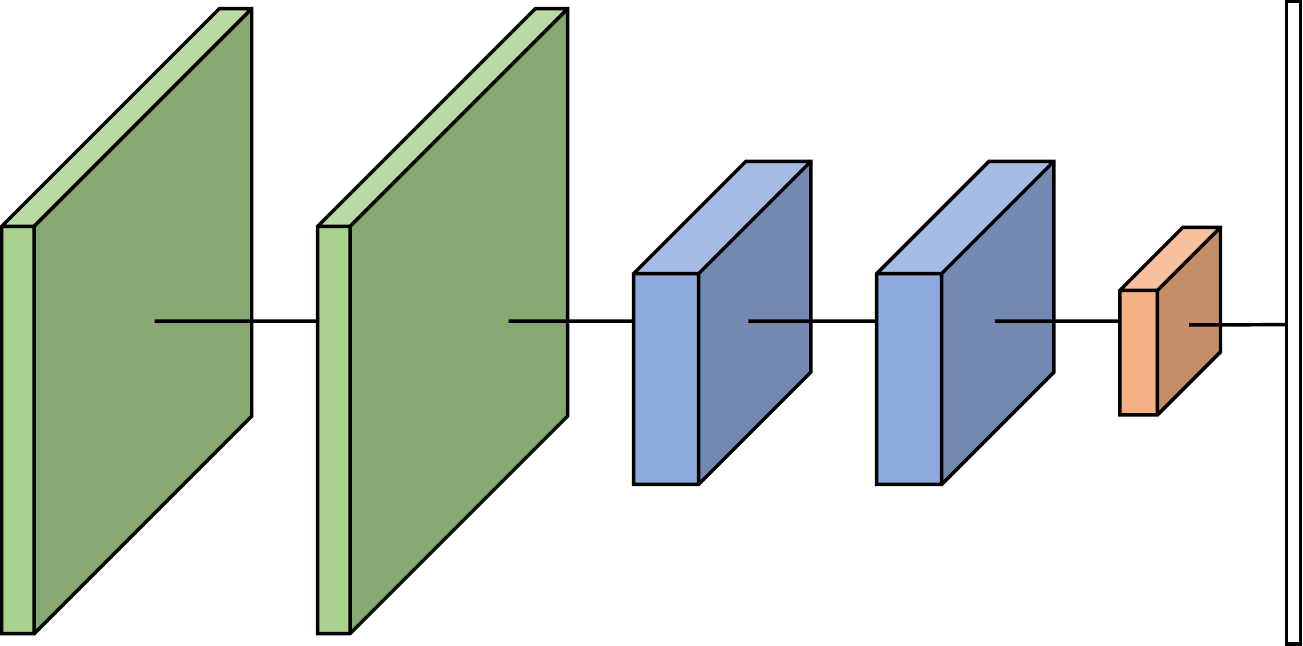} \\
\end{minipage}
\caption{Tiny nets.}
\end{figure*}

\subsection{Identification of connected components} 
We set a conservative arbitrary threshold of $0.001$ to detect duplicate images, and remove the edges of the k-nn graph that are above this threshold. We compute the connected components, and keep a single image per connected component. 

\begin{figure*}[t]
\raisebox{-.38\height}
{\includegraphics[width=0.5\linewidth]{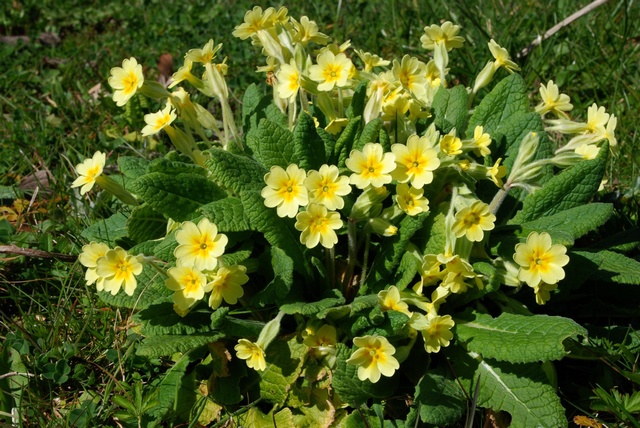}}
\hfill
\scalebox{0.6}{
\begin{tabular}{lp{8cm}}
n11610437 & bishop pine, bishop's pine, Pinus muricata \\
n11619455 & western larch, western tamarack, Oregon larch, Larix occidentalis \\
n11621281 & amabilis fir, white fir, Pacific silver fir, red silver fir, Christmas tree, Abies amabilis \\
n11626826 & red spruce, eastern spruce, yellow spruce, Picea rubens \\
n11710827 & cucumber tree, Magnolia acuminata \\
n11721642 & lesser spearwort, Ranunculus flammula \\
n11722342 & western buttercup, Ranunculus occidentalis \\
n11722621 & cursed crowfoot, celery-leaved buttercup, Ranunculus sceleratus \\
n11753562 & buffalo clover, Trifolium reflexum, Trifolium stoloniferum \\
n11840476 & desert four o'clock, Colorado four o'clock, maravilla, Mirabilis multiflora \\
n11874081 & yellow rocket, rockcress, rocket cress, Barbarea vulgaris, Sisymbrium barbarea \\
n11882426 & crinkleroot, crinkle-root, crinkle root, pepper root, toothwort, Cardamine diphylla, Dentaria diphylla \\
n11887750 & western wall flower, Erysimum asperum, Cheiranthus asperus, Erysimum arkansanum \\
n11889205 & tansy-leaved rocket, Hugueninia tanacetifolia, Sisymbrium tanacetifolia \\
... & ... \\
\end{tabular}}
\caption{\label{fig:flowers}
	Image that appears in the largest number of duplicate versions in \imnetbig (72), with a few of the corresponding synsets. 
}
\end{figure*}

For \imnetbig, the largest connected components are error images returned by image banks like Flickr for missing entries. This is an artifact of how the dataset was downloaded. 
The largest non-trivial cluster from \imnetbig is the image of a flower in Figure~\ref{fig:flowers}, that appears in 72 different synsets.
There seems to be some disagreement on the species of this flower, along with plain bad annotations.

\subsection{Statistics}  
Table~\ref{tab:dupstats} shows some statistics on the duplicates identified by our simple approach. \imnetbig has 10.4~\% duplicate images. 
In addition to these duplicates, we removed 930,757 images that overlap with \imnet, which means that \imnet is not a subset of \imnetbig in this paper. 
Within \imnet, we found 1~\% duplicates, which seems small enough not to remove them.
For \tiny, we found 9.5~\% duplicates and removed them, leaving the dataset with $71,726,550$ unique images.

\begin{table}
\begin{center}
\caption{\label{tab:dupstats}
	Duplicate statistics for the datasets we use.
}
\begin{tabular}{l@{\hspace*{10mm}}r@{\hspace*{10mm}}r}
\hline
Dataset 				& \# images & \# groups \\
\hline 	
\imnetbig				& 14,197,087	& 12,720,164 \\
\imnettrain				& 1,281,167		& 1,267,936	\\
\tiny					& 79,302,017	& 71,726,550\\
\hline
\end{tabular}
\end{center}
\end{table}

\section{TinyNet architectures}
\label{app:tinynet}

This appendix describes the convolutional architecture employed in Section \ref{sec:tinynets} of the main paper to evaluate the capacity saturation and the influence of training parameters and choices on this capacity. 

The architectures includes from 3 convolutional layers for TinyNet1 to 4 for TinyNet2 and TinyNet3. The first convolutional layer is 5x5. Each convolutional layer is followed by a Rectifier Linear Unit activation. The fully connected layer of TinyNet3 is larger than TinyNet2. 

\section{Filters}

\begin{figure*}[h]
{\small
\begin{minipage}{0.32\linewidth}
\centering $n$=100k, no augmentation \\ 
\myfig{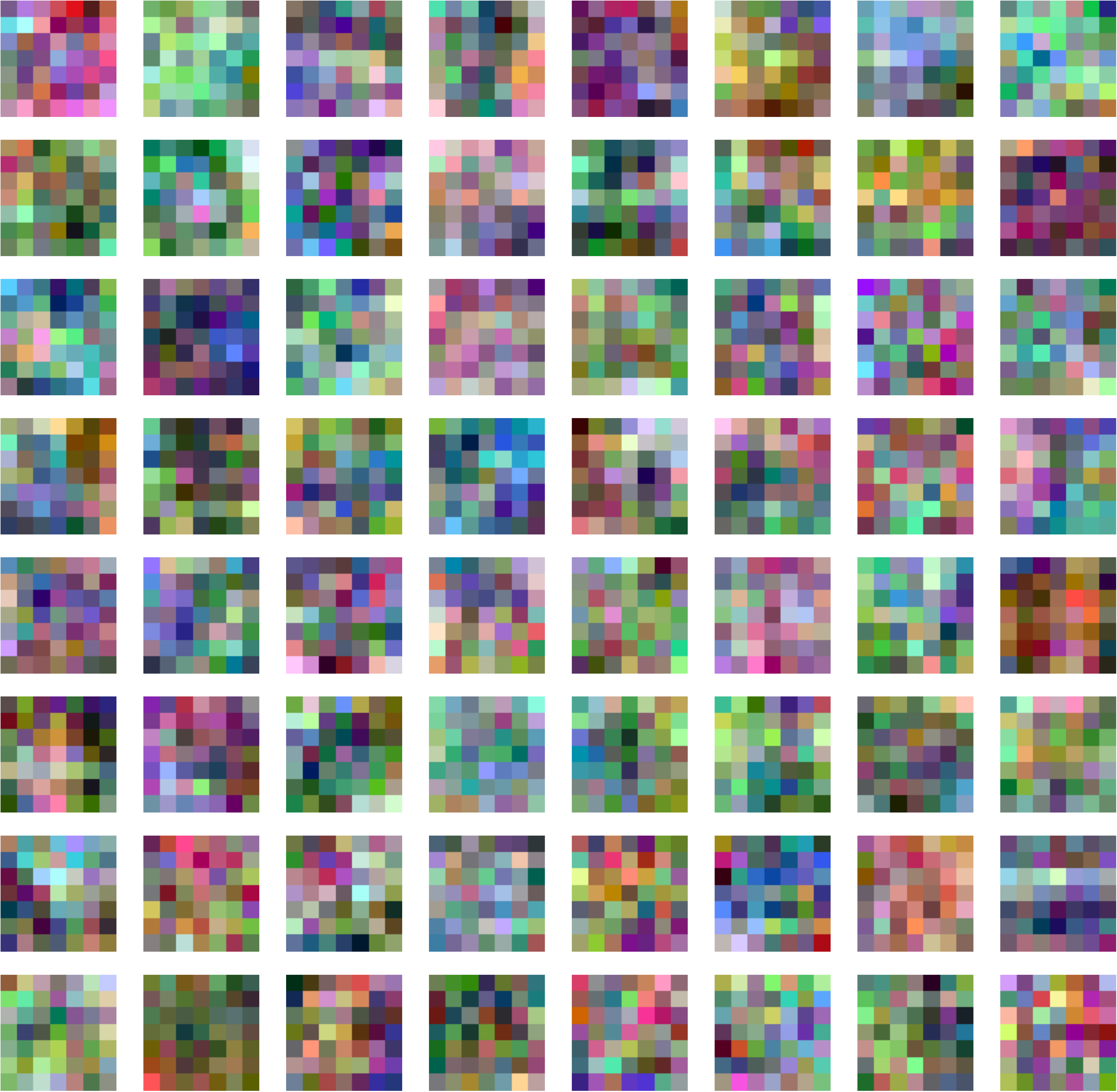} \\
\end{minipage}
\hfill
\begin{minipage}{0.32\linewidth}
\centering $n$=100k, flip \\ 
\myfig{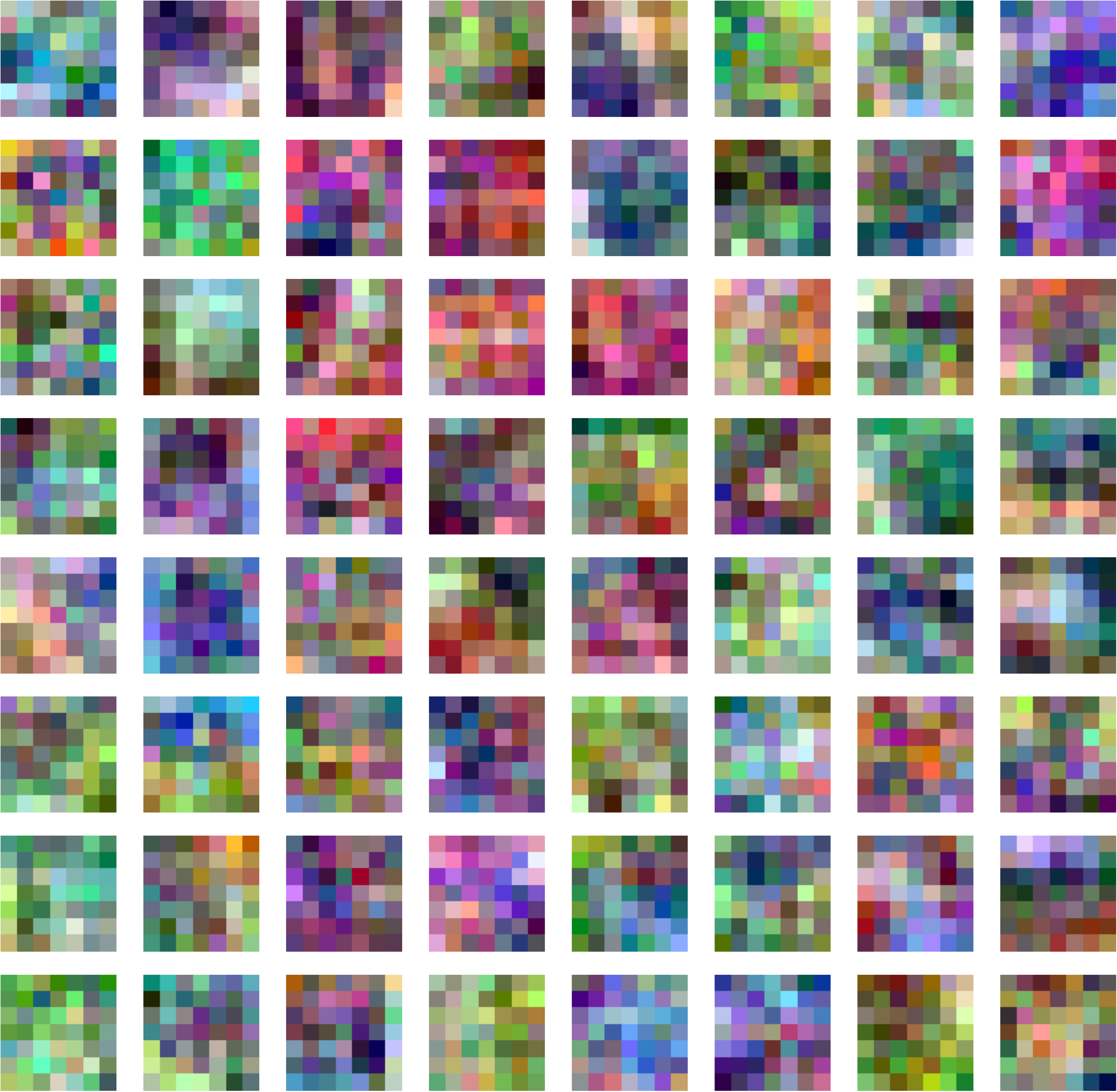} \\
\end{minipage}
\hfill
\begin{minipage}{0.32\linewidth}
\centering $n$=100k, flip+crop$\pm$2 \\ 
\myfig{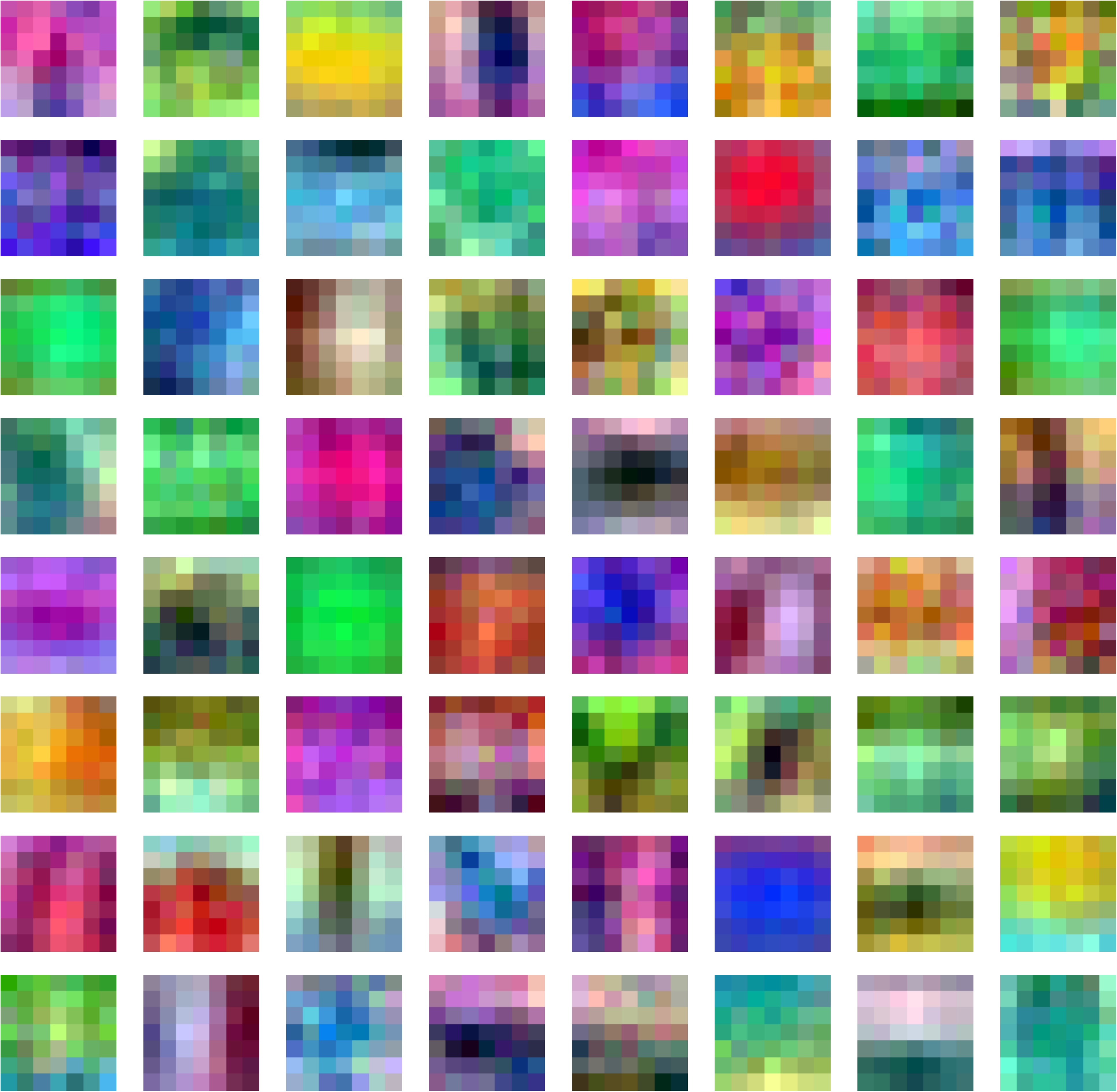} 
\end{minipage}}
\caption{Filters of the first convolutional layer (7x7, 64 filters) obtained when learning to explicitly memorize if an image was used for training or not. 
\label{fig:filters}}
\end{figure*}

The filters of the first convolutional layer are easy to visualize and contain interesting information on how the SGD optimized to the very first filter that is applied on the image pixels~\citep{krizhevsky2012imagenet,bojanowski2017unsupervised,paulin2017convolutional}. 
Figure~\ref{fig:filters} shows the filters obtained after training a Resnet-18. 
The filters for 10k images are very noisy compared to the smooth Gabor filters produced by supervised classifiers. 
This is probably due to the large capacity of the network, that is able to quickly overfit the data and does not need to update the filter weights beyond their random initialization. 
With more images, the filters become more uniform, exhibiting some specialization.
Interestingly, for $n$=100k with crop augmentation the filters have a clear uniform color. This is required for the output to be less sensitive to translations of up to 2 pixels. %

\section{Shadow models}
\label{app:shadow}

We evaluated the performance of shadow models on the partial-layers setting.
The setting is the following: we train $ 20 $ networks on the public dataset, each time holding out a different subset of images. 
For each network, we can thus compare the activations of train and held-out images.
These activations are not directly comparable between two different networks, because internal activations of a ReLU network have invariances (such as permutation of the neurons or positive rescaling).
To circumvent this issue, we learn a regression model that maps activations between two networks, and thus align activations of all the networks to the activations of the network under attack using the $ \ell_2 $ loss.
We then learn an attack model that predicts from the aligned activations whether the image was seen by the network at train time.

The results are shown in Table~\ref{tab:attack_shadow}. 
While performing better than random guessing, shadow models underperform the attack methods shown in Table~\ref{tab:attacks_partial}.
We believe that this is due to the complex processing involved in training shadow models on intermediate activations (notably the regression model), whereas the attacks of Section~\ref{sec:implicit} are more straightforward to train.

\begin{table}[t]
\centering
\begin{minipage}{\tabcaplen}
\caption{\label{tab:attack_shadow}
Accuracy of membership inference attacks before the softmax layer of the models (\textit{partial-layers}), using shadow models. 
}
\end{minipage}
\hfill
\begin{tabular}{@{}llcc@{}}
\toprule
Model     		& Augmentation 		& Attack accuracy \\ \midrule
Resnet101 	& None				& 60.6		 \\ 
			& Flip, Crop $\pm5$		& 61.4		 \\ 
			& Flip, Crop 			& 58.2		 \\ \midrule
VGG16     	& None   				& 73.8		  \\
    	 		& Flip, Crop $\pm5$		& 65.8		  \\
     			& Flip, Crop			& 55.2		  \\ \bottomrule
\end{tabular}
\end{table}

\end{appendices}

\end{document}